\documentclass[10pt,twocolumn,letterpaper]{article}

\usepackage{cvpr}              

\usepackage{graphicx}
\usepackage{amsmath}
\usepackage{amssymb}
\usepackage{booktabs}

\usepackage[pagebackref,breaklinks,colorlinks]{hyperref}

\usepackage[dvipsnames, table]{xcolor}
\usepackage{siunitx}
\mathchardef\mhyphen="2D
\usepackage{multirow}

\usepackage[capitalize]{cleveref}
\crefname{section}{Sec.}{Secs.}
\Crefname{section}{Section}{Sections}
\Crefname{table}{Table}{Tables}
\crefname{table}{Tab.}{Tabs.}

\def\cvprPaperID{4} 
\def\confName{PBVS@CVPR}
\def\confYear{2022}

\begin{document}

\title{Unsupervised Anomaly Detection from Time-of-Flight Depth Images}

\author{Pascal Schneider \href{https://orcid.org/0000-0002-0555-2694}{\includegraphics[scale=0.08]{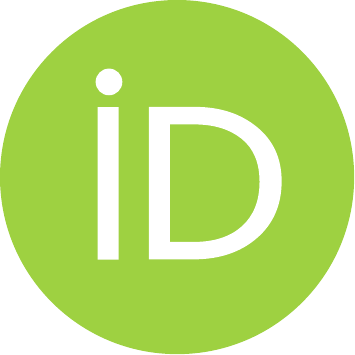}}, Jason Rambach, Bruno Mirbach, Didier Stricker\\
German Research Center for Artificial Intelligence (DFKI)\\
Trippstadter Str. 122, 67663 Kaiserslautern, Germany\\
{\tt\small firstname.lastname@dfki.de}}
\maketitle

\begin{abstract}
   Video anomaly detection (VAD) addresses the problem of automatically finding anomalous events in video data. The primary data modalities on which current VAD systems work on are monochrome or RGB images. Using depth data in this context instead is still hardly explored in spite of depth images being a popular choice in many other computer vision research areas and the increasing availability of inexpensive depth camera hardware. We evaluate the application of existing autoencoder-based methods on depth video and propose how the advantages of using depth data can be leveraged by integration into the loss function. Training is done unsupervised using normal sequences without need for any additional annotations. We show that depth allows easy extraction of auxiliary information for scene analysis in the form of a foreground mask and demonstrate its beneficial effect on the anomaly detection performance through evaluation on a large public dataset, for which we are also the first ones to present results on. 
\end{abstract}

\section{Introduction} \label{sec:intro}
Detecting anomalous events automatically from measurements of a system or the environment has been a long-standing field of research with a wide range of applications, such as network analysis \cite{ad_in_network_analysis} or the diagnosis of diseases from medical data \cite{ad_for_brain_mri}. Compared with other detection tasks, anomaly detection comes with some additional challenges, such as high heterogeneity of different anomalies and general uncertainty about their spatio-temporal properties \cite{dl_for_ad_review}. Video anomaly detection poses the additional problem of the high dimensionality of image data, which demands for methods that scale favourably in this respect. Deep neural networks have proven to be especially well suited in such cases across many different problems related to computer vision and also prevail in approaches to address video anomaly detection nowadays \cite{dl_for_vad_review}. Our work is also in line with this direction and employs an autoencoder-based principle in which a representation of normality is learned and subsequently used to detect anomalies (see \Cref{subsec:approach_ae_vad}).

The predominant data modality in VAD research is RGB\footnote{When referring to RGB video in the following, we also mean monochrome video (\eg by a surveillance camera) since it can be considered a special case of RGB.} video, since RGB cameras are the most common hardware to record data for image-based monitoring. Depth video, however, is yet hardly explored as an alternative, despite its popularity in many other fields of computer vision research \cite{hand_gr_depth, ppl_counting_depth}. Especially time-of-flight (ToF) camera technology has recently become increasingly affordable and improvements in recent hardware allow for high spatial resolution and range in measurements.

A general benefit of depth data is that it allows to detect and classify persons and objects in the scene, and to  localize them precisely in space, independently of the texture, color or light conditions. Because of this robustness, highly accurate people counting systems on the market are already based on ToF technology \cite{Irisys_people-counting,IEE_smartbuilding}. Research which explores the potential of recent time-of-flight cameras for more advanced and challenging applications as VAD is therefore clearly warranted.

Another advantage of using depth instead of RGB images is linked to the privacy aspect. Time-of-flight depth images generally provide little information useful for identifying a specific person,  This would also allow collecting raw data during operation to enlarge the training while preserving privacy. Overall, depth-based monitoring can help to provide safety where a monitoring system is mandatory and at the same time avoid the risk of negative societal impact that the implementation of such systems might entail.

Our work aims at advancing the newly emerging research in VAD on ToF depth data by investigating how existing algorithms could be adapted to this data modality in order to enhance detection performance. To summarize, the main contributions of our work are:
\begin{itemize}
    \item Performing generic unsupervised anomaly detection using a Time-of-flight depth image as input, accompanied by experiments validating the benefits of this modality compared to infrared intensity images.
    \item Presenting the first extensive evaluation on the newly introduced TIMo dataset \cite{schneider2021timo} over different anomaly detection methods and different anomaly cases.
    \item A combined approach for anomaly detection using depth images that outperforms other autoencoder methods proposed for RGB-based VAD by utilizing the advantages of depth data in foreground-aware loss functions.
    \item Introducing \emph{Vision Transformer}-based (ViT) autoencoder networks for the problem of anomaly detection and comparing their performance to convolution-based networks.
\end{itemize}

The rest of the paper is structured as follows: \Cref{sec:rel_work} gives a short overview of other methods and how they relate to the one we propose. We will then present our approach in \Cref{sec:approach} and the results of the evaluation on the TIMo dataset and a discussion thereof in \Cref{sec:exp}. We conclude our work and contributions in \Cref{sec:conc}.

\section{Related Work} \label{sec:rel_work}
The general task of anomaly detection has been a research topic for a long time and many different approaches were presented over the time \cite{outlier_analysis_book}. With the great successes of deep learning in pattern recognition-related task, deep neural networks have also been increasingly adopted for anomaly detection \cite{dl_for_ad_review, dl_for_vad_review, vad_overview_semisup_unsup}.

The subtopic of video anomaly detection first gained widespread attention in the mid- to late-2000s. Many of the approaches that were proposed during this period used optical flow or object trajectories to represent motion and stochastic models such as \emph{Hidden Markov Models} to capture scene activity \cite{vad_review_2012, unusual_event_detection_2008}. 

Since VAD can be thought of as a frame-level binary classification task, one way of approaching the problem is to apply supervised learning techniques. A concept that has proven to be useful in this context for anomaly detection is \emph{contrastive loss}. The general idea is to apply the notion of \emph{metric learning} in order to train a DNN to create embeddings of the samples such that normal ones form clusters while anomalous ones are more distant in the embedding space. Examples of works employing contrastive loss are those by Köpüklü \etal \cite{driver_anomaly_dataset} and Khan \etal \cite{modified_contr_learning_driving_ad}. These works are moreover both among the very few works as of now that use depth video as the primary data modality, but differ from our work in that we aim at an unsupervised approach.

A key challenge for many anomaly detection applications is the rarity of anomalous events. Creating large datasets of anomalies can thus often turn out to be impractical, since it is inherently an \emph{open set recognition problem}. The lack of knowledge about what kind of anomalies might occur during inference in the real world makes it impossible to compile even a remotely exhaustive training set. Consequently, unsupervised approaches are especially interesting in this context. Instead of learning to distinguish between normal and anomalous samples directly, these methods usually aim at learning a representation of patterns of normal motion and spatial appearance and some mechanism to detect anomalies as significant deviations thereof. 

Our work is based on the general approach of training autoencoder networks to perform either reconstruction or prediction of video frames, which is detailed in \Cref{subsec:approach_ae_vad}. There are a few other concepts to approach anomaly detection in an unsupervised way. These include using \emph{generative adversarial networks} (GAN) \cite{image_ad_with_gan} and clustering or estimating densities in the latent space of an autoencoder \cite{dl_for_ad_review}. These methods come with their own challenges, though. Getting a GAN to train successfully is known to often be difficult. Clustering or estimating densities on the other hand requires the latent space to be designed with a dimensionality that provides a good tradeoff between representational capacity and still allow the algorithm for clustering/density estimation to run with practicable computational cost. The autoencoder-based reconstruction and prediction concept that we employ tends to be rather straightforward in comparison.

An advantage of using RGB video lies in the availability of many well-established algorithms that can be used to extract auxiliary information to aid the anomaly detection. An example is the frequent use of optical flow to help detecting motion-based anomalies \cite{shanghai_campus_dataset_and_algo, any_shot_seq_ad, vad_with_of_and_cae, visual_ad_tem}. Some works use much higher-level information, \eg human skeleton estimations \cite{skeleton_regularity_ad} or object detections \cite{self_sup_multi_task_vad, object_centric_ae_ad, any_shot_seq_ad}. A potential issue that can arise from using information such as human skeletons or bounding boxes of detected objects is the risk that the algorithms used for their estimation might not work reliably in case of anomalies. Many of these algorithms implicitly make use of the \emph{\iid assumption}, which is likely to be violated by the data produced by an anomalous event. Moreover, many of these approaches do not translate well to ToF depth video directly because the algorithms are not as mature yet or the concepts themselves are harder to apply to it. \eg, optical flow computation on ToF data suffers from lack of texture and regions of invalid pixels (for a description of ToF camera artefacts see, e.g. \cite{Kinect_Azure,Kinect_eval}).

The use of foreground masks is a key aspect of our work. The basic idea itself is not new, an example of how it has been successfully applied in the training of a neural network is the work of Song \etal \cite{mask_guided_pr}. They proposed using binary mask to improve person re-identification and demonstrated the positive effect on performance. Their approach to generating the masks is based on fully connected network trained on labeled data. Since depth data facilitates foreground segmentation significantly compared to RGB, we can employ a more efficient and versatile method to generate the masks, which is detailed in \Cref{sec:fg_mask_gen}.

\section{Approach} \label{sec:approach}
The networks we use in our evaluations employ the principle of learning latent representations of normality by training an autoencoder in a unsupverised way. This principle is outlined below and we subsequently detail the specific network architectures.

\subsection{Autoencoder-based VAD} \label{subsec:approach_ae_vad}
A popular choice when approaching anomaly detection in an unsupervised way are mechanisms based on autoencoder networks and proxy tasks. Autoencoder architectures are generally designed in such a way that they compress a high-dimensional input into a low-dimensional latent representation (\ie the \emph{encoder}) and subsequently use this representation to reconstruct the original input data (\ie the \emph{decoder}). The reconstruction is a proxy task, since the relevant output is not the reconstruction itself, but its deviation from the ground truth, which is captured by a loss function and governs the training process of the network. This basic principle is used in video anomaly detection to have the autoencoder learn a latent representation of normality by providing it a large number of normal samples to train with. Normal parts of a video are thus expected to be reconstructed with low error because of their similarity to what the network has learned to represent. Anomalies are in turn expected to cause a higher loss when being reconstructed. The loss can hence be interpreted as an anomaly score. An alternative proxy task is the prediction of future frames. The same logic applies, normal events are expected to be predicted more precisely while anomalous events are harder to predict and thus lead to higher loss.

\subsection{Network Architectures}
We evaluated a total of five different autoencoder-based networks. All of them are based on the concept of using either reconstruction loss or prediction loss as an anomaly score. Some existing approaches reconstruct or predict whole sequences of frames (\eg \cite{conv_ae_ad}), which means that some form of interpolation technique is necessary to obtain frame-level anomaly scores. We constrained the set of approaches to such ones that produce an anomaly score that is directly associated to an individual frame in order to facilitate the comparison. This is achieved by having all networks only predict or reconstruct a single frame.

For the convolutional networks, we consistently use a kernel size of $5{\times}5$. We do not employ regularization techniques such as dropout. We only added slight normally distributed noise to the images before it is processed by the autoencoder, which is a common technique to facilitate that the autoencoder learns more robust latent representations \cite{denoising_ae}. This is moreover independent from the specific network architecture and can thus be easily applied to all the ones we present. The sequences for the prediction-based networks each consist of \num{4} frames.

\subsubsection{Reconstruction Convolutional Autoencoder (R-CAE)}
The first network architecture we investigate is a convolutional autoencoder, which performs reconstruction of single frames. This basic principle was already used in early deep learning-based video anomaly detection, such as in the work by Hasan \etal \cite{conv_ae_ad}.

The network consists of three convolutional layers in the encoder with subsequent $2{\times}2$ max-pooling operations. The number of filters is reduced from \num{32} at the input layer down to \num{8} in the central latent space and then increased again to \num{64} through transposed convolutional layers and upsampling operations until the final convolution layer performs the image reconstruction step.

\subsection{Prediction Convolutional Autoencoder (P-CAE)}
This network also uses the basic principle behind convolutional autoencoders, but instead performs prediction. The sequence of four images is presented to the network as a tensor where the images are ordered along an additional tensor dimension. Hasan \etal \cite{conv_ae_ad} proposed a network that is similar to the one presented here, but uses different kernel sizes, strides and reconstructs the whole input sequence instead of predicting a single future frame.

The network's encoder consists of five convolutional layers with subsequent $2{\times}2$ max-pooling operations. The decoder employs five transposed convolutional layers with $2{\times}2$ upsampling operations performed after each layer. The number of filters is reduced from initially \num{64} down to \num{4} in the central latent space and then increased again to \num{64} before the final layer, which is again a transposed convolutional layer that generates the frame prediction.

\subsection{Prediction Convolutional LSTM (P-ConvLSTM)}
Networks using the \emph{long short-term memory} (LSTM) \cite{lstm} concept usually involve flattening the data, which can be undesirable when applied to image data because spatial structure is partially lost this way. The \emph{ConvLSTM} architecture originally proposed by Shi \etal in \cite{convlstm_orig_paper} aims at addressing this problem by combining LSTM cells and the convolution operation. This architecture has already been successfully used in the VAD context by Luo \etal \cite{conv_lstm_ad}. Their network design is the basis for the one we propose. The network consists of \num{6} ConvLSTM cells with a number of \num{8} hidden dimensions in each cell.

\subsection{Reconstruction Vision Transformer (R-ViT-AE)}
Just like the R-CAE, this network is trained to reconstruct the frame it receives as an input. The difference lies in the internal architecture. Instead of convolutions, this network is based on the concept of a transformer applied to the image domain \cite{vision_transformer}, \ie a \emph{vision transformer} (ViT). The network we use builds on the work and accompanying implementation of Ahmed \etal \cite{self-supervised_vision_transformer}, who propose a self-supervised vision transformer architecture which they apply to the problem of reconstructing corrupted images. Our network is set up with \num{4} layers, a patch size of $16{\times}16$ pixels and an embedding dimension of \num{8}.

\subsection{Prediction Vision Transformer (P-ViT-AE)}
Following the same concepts as before, this network uses the same architecture as the R-ViT-AE but instead receives a sequence of the four previous frames and outputs a prediction of the next frame.

\subsection{Foreground Mask Generation} \label{sec:fg_mask_gen}
\begin{table}[tb]
  \centering
  \begin{tabular}{r|c|c|c|c|c|c}
    \toprule
    \textbf{Param.} & $K$ & $K_\mathrm{kinect}$ & $\Delta P_\mathrm{max}$ & $\alpha$ & $T_\mathrm{W}$ & $N_\mathrm{H}$ \\
    \midrule
    \textbf{Value}     & \num{1.25} & $5 \cdot 10^{-4}$       & \num{100}         & \num{0.4}     & \num{300}  & \num{90} \\
    \bottomrule
  \end{tabular}
  \caption{Choice of parameters for the algorithm of Braham \etal \cite{physically_based_BGS}}
  \label{tab:bgs_parameters}
\end{table}

\begin{figure}[tb]
  \centering
  \begin{subfigure}{0.5\linewidth}
    \includegraphics[width=1\linewidth]{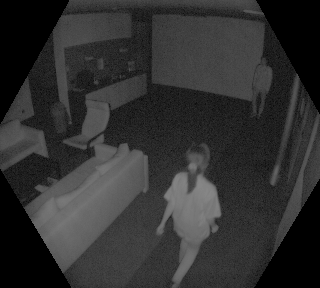}
    \includegraphics[width=1\linewidth]{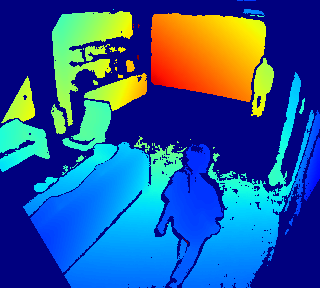}
    \includegraphics[width=1\linewidth]{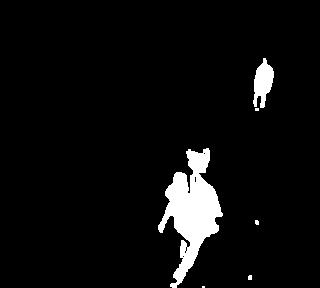}
    \caption{A0186, frame no. 76}
    \label{fig:fg_mask_example_a}
  \end{subfigure}
  \begin{subfigure}{0.45\linewidth}
    \includegraphics[width=1\linewidth]{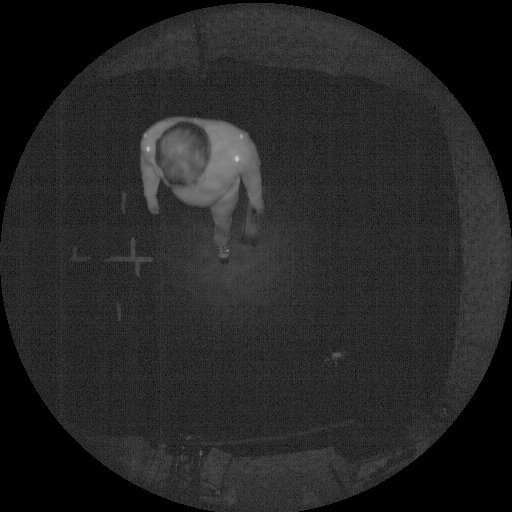}
    \includegraphics[width=1\linewidth]{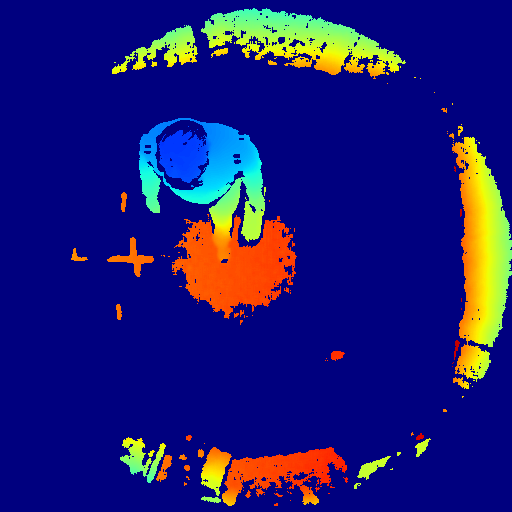}
    \includegraphics[width=1\linewidth]{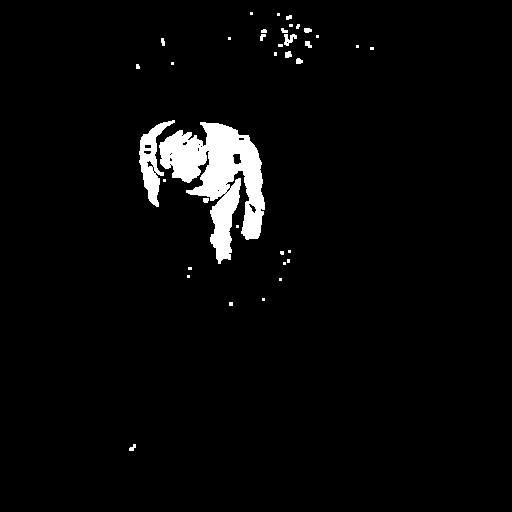}
    \caption{A1454, frame no. 94}
    \label{fig:fg_mask_example_b}
  \end{subfigure}
  \caption{Examples of the computed foreground masks for two selected frames. Depth is visualized using a color mapping where depth increases from blue to red. Dark blue corresponds to invalid measurements. (Top to bottom: Infrared, depth, foreground mask.)}
  \label{fig:fg_mask_examples}
\end{figure}

Among the most basic information that can be used to analyze a scene based on images is the segmentation into foreground and background. Hence, this problem has been the subject of long-standing research in computer vision research. Our motivation for applying this segmentation in the context of anomaly detection is the notion that anomalous events occur in the foreground of an observed scene.

For RGB images, background segmentation turns out to be challenging in many cases \cite{bgs_review}. When using depth images on the other hand, the task of segmenting foreground and background becomes much more straightforward and can be achieved with relatively simple algorithms and low computational effort. We use the algorithm presented by Braham \etal in \cite{physically_based_BGS}. It is pixel-based and the parameters directly relate to physical quantities, which allows setting parameters based on knowledge of the sensor and basic assumptions about the scene (\eg a rough order of magnitude estimate of how fast objects in the scene move). It is therefore well-suited to be part of an unsupervised approach. The parameters we used are given in \Cref{tab:bgs_parameters}.

For a detailed description of the algorithm, we refer to \cite{physically_based_BGS}. The basic working principle is that each pixel maintains a simple model of its historical state and can switch between being considered background or foreground. Such a transition might occur for instance if a background pixel has a sudden large change in its current depth value. This indicates that the pixel is now part of the image's foreground -- as opposed to being background where only small changes due to noise would be expected. Invalid pixel states are also considered, which is important for ToF data with potentially lots of invalid measurements due to \eg flying pixels. The algorithm is bootstrapped by defining the initial state as being all background, which corresponds to an empty scene. \Cref{fig:fg_mask_examples} shows example of masks that were generated by our implementation of the algorithm.

The authors of \cite{physically_based_BGS} propose applying a median filter as a post-processing step to the image to mitigate noise. We observed that this leads to loss of some finer details in the mask in our case and consequently do not perform median filtering. We instead apply a Gaussian filter with a kernel size of $5{\times}5$. The foreground mask is thus not binary and can extend slightly into regions around the foreground object's valid pixels. The borders of flying pixel around objects often also carry some shape information, which can be partly captured this way.

\subsection{Loss Functions} \label{sec:loss_fns}
We trained all of the networks with three different loss functions to evaluate their effect on the anomaly detection performance. The loss functions are detailed below.

\subsubsection{MSE}
The MSE loss we employ is the customary \emph{mean squared error} function, sometimes also referred to as the squared L2 norm. The target used to compute the output's error is the subsequent frame in case of the prediction and the current frame in case of reconstruction.

\subsubsection{F-MSE}
The \emph{foreground MSE} (F-MSE) applies the foreground mask values directly to the MSE computation. \Cref{eq:fmse} details the computation of the F-MSE loss over $n$ output pixel values $\hat{y}_i$ and target pixel values $y_i$ and the value of the foreground mask $m_i$ at the respective position. Background regions where $m_i = 0$ thus do not increase the overall loss.

\begin{equation} \label{eq:fmse}
    \mathrm{F\mhyphen MSE}\,(y, \hat{y}) = \frac{\sum_{i=1}^n m_i \, (y_i - \hat{y}_i)^2}{n}
\end{equation}

\subsubsection{W-MSE}
The \emph{weighted MSE} (W-MSE) also computes the MSE between the output and the target image, but adds a factor to put stronger weight on error inside the foreground mask. The choice of a weighting factor of course allows for infinitely many different choices. We chose a factor of \num{8}, \ie the loss for pixels within the foreground mask is increased by adding eight times the loss times the mask value at the position to the MSE. The background loss is thus still relevant for the overall loss, but there is a strong emphasis on the foreground. The value \num{8} for the weighting factor was selected using a rough grid search in the range of $[2, 128]$ and lead to distinguishable results of the W-MSE loss compared to MSE and F-MSE. The W-MSE loss can be expressed as a combination of the MSE and F-MSE functions as shown in \Cref{eq:wmse}.

\begin{equation} \label{eq:wmse}
    \mathrm{W\mhyphen MSE}\,(y, \hat{y}) = \mathrm{MSE}\,(y, \hat{y}) + 8 \cdot \mathrm{F\mhyphen MSE}\,(y, \hat{y})
\end{equation}

\subsection{Post-Processing the Anomaly Score} \label{sec:post_proc}
The training process does not enforce temporal smoothness of the anomaly score, which can lead to fluctuation with high amplitude across time. This is a known issue in anomaly detection research and can be addressed by employing a low-pass filtering technique to post-process the anomaly score \cite{driver_anomaly_dataset}. We use a moving average operation with a sliding-window of length of \num{10}. This attenuates the fluctuations in the anomaly score and also leads to a slight improvement in the overall detection performance (see supplementary material for more details).

\section{Experiments and Analysis} \label{sec:exp}
All five networks were evaluated in combination with each of the three loss functions. The optimizer we use is \emph{Adam} with a learning rate of \num{0.001}. During the development of the network architectures, we observed that the loss already converges during the first epoch and the anomaly detection performance \wrt the metrics we use did not change significantly when training for longer. We thus trained the networks only for a single epoch and leave the learning rate unchanged during the training. Training took place on a single \emph{Nvidia GeForce GTX Titan X}. Details on inference time performance can be found in the supplementary material.

\begin{figure*}[tb]
  \centering
  \begin{subfigure}{0.245\linewidth}
    \includegraphics[width=1\linewidth]{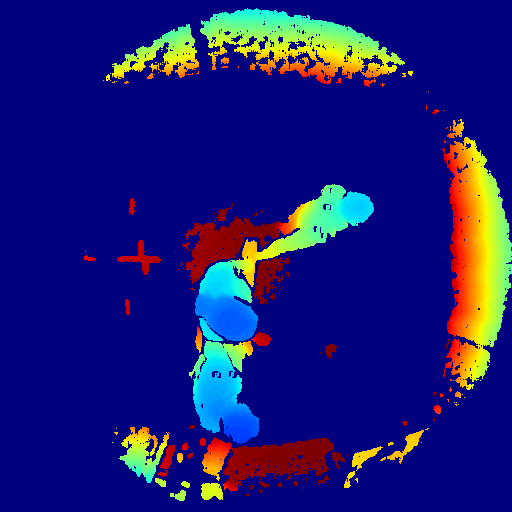}
    \caption{Ground truth}
    \label{fig:example_prediction_GT}
  \end{subfigure}
  \begin{subfigure}{0.245\linewidth}
    \includegraphics[width=1\linewidth]{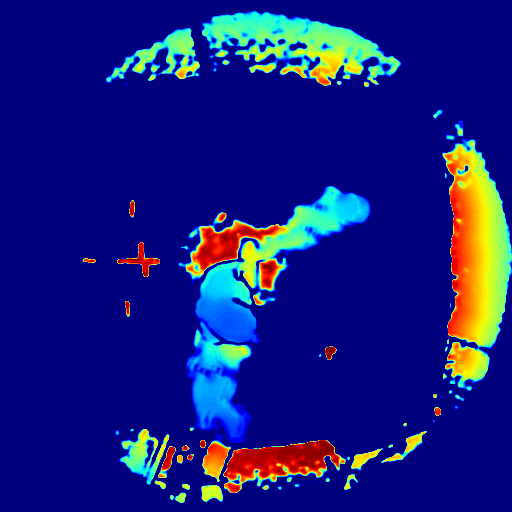}
    \caption{Prediction with MSE loss}
    \label{fig:example_prediction_MSE}
  \end{subfigure}
    \begin{subfigure}{0.245\linewidth}
    \includegraphics[width=1\linewidth]{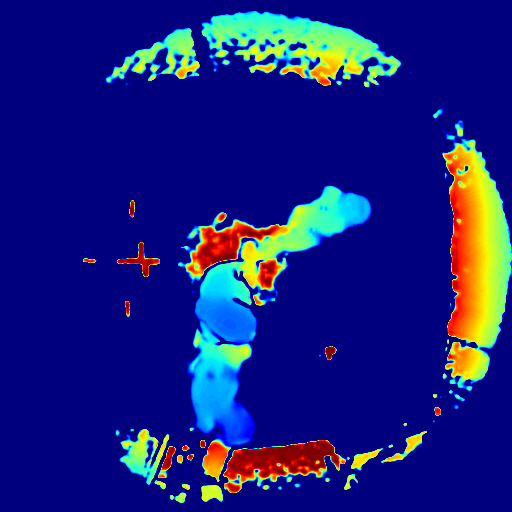}
    \caption{Prediction with W-MSE loss}
    \label{fig:example_prediction_WMSE}
  \end{subfigure}
    \begin{subfigure}{0.245\linewidth}
    \includegraphics[width=1\linewidth]{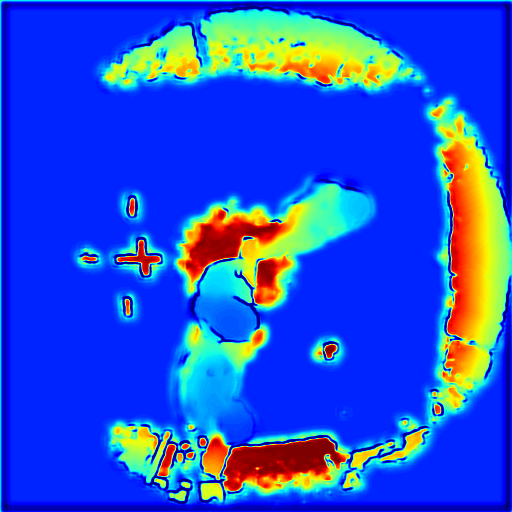}
    \caption{Prediction with F-MSE loss}
    \label{fig:example_prediction_OFMSE}
  \end{subfigure}
  \caption{Example of the prediction of a frame predicted by the ConvLSTM network trained with different loss functions.}
  \label{fig:example_predictions}
\end{figure*}

\subsection{Dataset and Data Normalization} \label{sec:dataset}
We use the TIMo dataset \cite{schneider2021timo} for evaluation, which features depth and infrared videos of normal and anomalous human behaviour in two indoor scenes captured by a \emph{Microsoft Azure Kinect} ToF camera. One of the scenes uses a \emph{tilted view} perspective and the camera's narrow field-of-view configuration, the other one uses a \emph{top-down view} and a wide field-of-view. The data from these two setups thus differs in several aspects including image size and aspect ratio and we evaluated the methods on each of these parts of the dataset separately.

To the best of our knowledge, TIMo is currently the only public large-scale dataset for unsupervised anomaly detection on depth videos. The dataset was released under the CC BY-NC-SA 4.0 license \footnote{\url{https://vizta-tof.kl.dfki.de/timo-dataset-overview/}}.

The anomalies in the test set are annotated in the form of the index of the first and last anomalous frame of the anomaly, which can be converted to a binary label about a frame being normal or anomalous. The error of the reconstruction or prediction from the autoencoder can in principle also be used to attempt a spatial localization of anomalies by employing a heatmap of the loss. However, since TIMo does not feature spatial annotations about anomalies, we only evaluate the temporal aspect.

The depth images are normalized to the $[0,1]$ range with a simple linear min-max normalization with the minimum set to \SI{0}{\meter}. The maximum is set to \SI{3.5}{\meter} for the top-down data and to \SI{11}{\meter} for the tilted-view data. These values roughly correspond to the maximum depth that can occur in the measurements given each scene's geometry.

We also ran experiments using the infrared amplitude images from the dataset, which were recorded simultaneously by the Microsoft Azure Kinect. The IR amplitude images have a high dynamic range. This is due to strong differences in the remission properties of surfaces and the fact that the camera uses active IR illumination, which leads to objects closer to the camera causing much higher IR amplitude in the image. A simple min-max normalization is therefore less suitable in this case. We instead reduce the dynamic range with a logarithmic mapping:
\begin{equation}
    x' = \frac{\log (1+x)}{2^{16} - 1}
\end{equation}
Examples of how the IR images appear after this mapping can be seen in \Cref{fig:fg_mask_examples}.

\subsection{Metric}
The task of frame-level video anomaly detection can be viewed as a binary classification of each frame as either normal or anomalous. However, the output of our networks is an anomaly score which has to be compared to a threshold to convert it to a binary label. It is therefore common practice in VAD research to report the results in terms of the area under the ROC curve (AUC-ROC) instead of recall and precision for a specific threshold. 

For the first four frames of each sequence, the prediction-based networks cannot generate a prediction due to the lack of a full sequence of previous frames. We therefore skip these frames and no anomaly score is produced for them. In order to evaluate all approaches on the exact same data, we also leave out these frames in the evaluation of the reconstruction-based networks. Since sequences start with an empty scene and typically consist of hundreds of frames, the effect of omitting these first frames has very little impact on the results.

\subsection{Results on Depth Data}
\begin{table*}[tb]
\centering
\begin{tabular}{c c|c|c|c||c|c|c|} \cline{3-8}
                                 & & \multicolumn{3}{c||}{Tilted View}    & \multicolumn{3}{c|}{Top-Down View} \\ \cline{3-8}
                                 & & MSE  & F-MSE   & W-MSE &               MSE  & F-MSE & W-MSE \\ \hline 
\multicolumn{1}{|c}{\multirow{5}{*}{\rotatebox[origin=c]{90}{ \textbf{Depth} }}} & \multicolumn{1}{|c|}{R-CAE}      & 66.4 & 68.5 & \textbf{70.0}     &      56.4 & \textbf{73.2}  & 63.9  \\
\multicolumn{1}{|c}{} & \multicolumn{1}{|c|}{P-CAE}      & 71.4 & 68.6 & \textbf{78.1}     &      52.1 & 66.0  & \textbf{67.8}  \\
\multicolumn{1}{|c}{} & \multicolumn{1}{|c|}{R-ViT-AE}   & 64.9 & 71.2 & \textbf{71.7}     &      57.0 & \textbf{65.3}  & 61.5  \\
\multicolumn{1}{|c}{} & \multicolumn{1}{|c|}{P-ViT-AE}   & 65.1 & \textbf{71.2} & 70.5     &      57.3 & \textbf{63.0}  & 60.3  \\
\multicolumn{1}{|c}{} & \multicolumn{1}{|c|}{P-ConvLSTM}  & 62.8 & \textbf{67.5} & 67.1    &      55.3 & \textbf{65.6}  & 59.1  \\ \hline \hline
\multicolumn{1}{|c}{\multirow{5}{*}{\rotatebox[origin=c]{90}{ \textbf{Infrared} }}} & \multicolumn{1}{|c|}{R-CAE}      & 62.6 & 64.2 & \textbf{65.5} & 56.5 & \textbf{60.2} & 58.7  \\
\multicolumn{1}{|c}{} & \multicolumn{1}{|c|}{P-CAE}      & 61.2 & 63.8 & \textbf{64.9} &	50.9 & \textbf{57.7} & 55.9  \\
\multicolumn{1}{|c}{} & \multicolumn{1}{|c|}{R-ViT-AE}   & 56.2	& \textbf{64.5} & 63.0 & 49.4 & \textbf{57.7} & 56.3  \\
\multicolumn{1}{|c}{} & \multicolumn{1}{|c|}{P-ViT-AE}   & 56.2 & \textbf{64.2} & 62.3 & 49.4 & \textbf{57.1} & 54.3  \\
\multicolumn{1}{|c}{} & \multicolumn{1}{|c|}{P-ConvLSTM} & 62.2 & \textbf{63.8} & 63.6 & 49.7 & \textbf{57.9} & 56.2  \\ \hline
\end{tabular}
\caption{Frame-level AUC (\%) for the different networks trained and evaluated with each of the three loss functions. The best result of each network on each part of the dataset is marked in bold.}
\label{tab:main_results}
\end{table*}

\begin{table*}[tb]
	\centering
	\begin{tabular}{c c|c|c|c||c|c|c||c|c|c|} \cline{3-11}
		& & \multicolumn{3}{c||}{Aggressive Behavior}    & \multicolumn{3}{c|}{Medical Issue} & \multicolumn{3}{c|}{Left-behind Object}\\ \cline{3-11}
		                                                                                       &                                  & MSE   & F-MSE         & W-MSE         & MSE  & F-MSE         & W-MSE         & MSE  & F-MSE         & W-MSE                 \\ \hline
		\multicolumn{1}{|c}{\multirow{5}{*}{\rotatebox[origin=c]{90}{ \textbf{Tilted View} }}} & \multicolumn{1}{|c|}{R-CAE}      & 76.78 & 78.4          & 82.3          & 48.0 & 52.3          & 49.5          & 66.6 & 65.2          & 68.6                  \\
		\multicolumn{1}{|c}{}                                                                  & \multicolumn{1}{|c|}{P-CAE}      & 79.3  & 78.7          & \textbf{85.9} & 59.9 & 52.3          & \textbf{65.3} & 73.4 & 66.8          & \textbf{79.1}         \\
		\multicolumn{1}{|c}{}                                                                  & \multicolumn{1}{|c|}{R-ViT-AE}   & 68.3  & 76.4          & 79.3          & 53.2 & 60.3          & 58.1          & 71.8 & 69.6          & 73.7                  \\
		\multicolumn{1}{|c}{}                                                                  & \multicolumn{1}{|c|}{P-ViT-AE}   & 68.5  & 75.9          & 79.3          & 53.6 & 61.8          & 55.4          & 72.6 & 70.2          & 72.2                  \\
		\multicolumn{1}{|c}{}                                                                  & \multicolumn{1}{|c|}{P-ConvLSTM} & 68.9  & 78.6          & 76.3          & 50.9 & 52.3          & 52.6          & 64.9 & 64.5          & 66.7                  \\ \hline \hline
		\multicolumn{1}{|c}{\multirow{5}{*}{\rotatebox[origin=c]{90}{ \textbf{Top-Down V.} }}} & \multicolumn{1}{|c|}{R-CAE}      & 61.7  & 86.9          & 76.3          & 54.9 & \textbf{75.6} & 64.8          & 56.1 & \textbf{73.2} & 63.3                  \\
		\multicolumn{1}{|c}{}                                                                  & \multicolumn{1}{|c|}{P-CAE}      & 35.9  & \textbf{93.0} & 82.7          & 55.2 & 69.9          & 73.8          & 52.2 & 64.6          & 65.1                  \\
		\multicolumn{1}{|c}{}                                                                  & \multicolumn{1}{|c|}{R-ViT-AE}   & 61.5  & 91.1          & 69.4          & 57.8 & 68.4          & 62.3          & 56.1 & 64.3          & 61.6                  \\
		\multicolumn{1}{|c}{}                                                                  & \multicolumn{1}{|c|}{P-ViT-AE}   & 65.2  & 92.5          & 75.0          & 58.2 & 65.9          & 60.8          & 56.2 & 62.1          & 60.2                  \\
		\multicolumn{1}{|c}{}                                                                  & \multicolumn{1}{|c|}{P-ConvLSTM} & 74.2  & 91.6          & 82.9          & 54.0 & 67.1          & 57.8          & 54.6 & 65.5          & 59.1                  \\ \hline
	\end{tabular}
	\caption{Frame-level AUC (\%) for three categories consisting of different types of anomalies from the TIMo dataset. The best result for each anomaly category of each part of the dataset is marked in bold.}
	\label{tab:anomaly_group_results}
\end{table*}
The results of the evaluation for the depth data from the TIMo dataset are shown in the upper part of \Cref{tab:main_results}. The beneficial effect on anomaly detection performance when using a foreground-aware loss function is clearly visible from this data. The only outlier in this respect is the P-CAE network, which undergoes a slight decrease in performance on the tilted view data in combination with the F-MSE loss function.

Another clear finding are the overall differences in performance on the tilted view and top-down view data. With performances not far off the mark of \SI{50}{\percent} AUC, all five networks perform poorly on the top-down data when using the MSE loss, but also profit more from switching to a foreground-aware loss compared to the results on tilted view data.

The effect of the loss functions on the reconstructions and predictions is also in part directly visible. \Cref{fig:example_predictions} shows an example of a prediction from the P-ConvLSTM network in combination with each of the three loss functions. It can be seen that the background is reconstructed with much less detail by the network trained with F-MSE loss. Edges in the background appear blurred, indicating that the internal cutoff frequency which the latent representation of the background is able to achieve is lower than for the other networks. The values of invalid pixel regions -- visualized as dark blue in \Cref{fig:example_predictions} -- are also only reconstructed with a significant bias. The background being reconstructed at all is likely to be caused by noise in the background mask (see \Cref{fig:fg_mask_examples}), which allows for some background-related loss to affect training even when using the F-MSE loss function.

The effect of putting more weight on foreground regions also becomes visible at close inspection of the predictions in \Cref{fig:example_prediction_MSE} (MSE) and \Cref{fig:example_prediction_WMSE} (W-MSE). The network trained with W-MSE features more detail in its prediction of the people in the scene. 

The comparison of the anomaly detection performance achieved by the vision transformer-based and the convolution-based networks shows no clear superiority of one approach over the other. The R-ViT-AE and the P-ViT-AE network show a slightly larger difference in performance between using the infrared and depth images. 

\subsection{Results on Infrared Images (IR)}
The IR images from the TIMo dataset are structurally very similar to the depth images in that they have the same image size, only have a single-channel, and are moreover time-synchronized with the depth images. This allows training the networks we present without any alterations on IR images compared to depth to evaluate the effect of this switch in data modality directly.

The results on the infrared data confirm the positive effect of using foreground masks to compute the reconstruction or prediction loss. The lower part of \Cref{tab:main_results} shows the results of the evaluation. Models trained with loss functions featuring the foreground masking again consistently outperform the ones trained with MSE loss, this time without exception. However, the overall performance when using IR instead of depth decreases. The average AUC across all combinations of network architectures, loss functions and parts of the dataset is \SI{63.5}{\percent} for depth and \SI{58.9}{\percent} for IR.

Jointly using IR and depth data to run the anomaly detection on a combination of both was not considered in this work, but is a possible subject to future research.

\begin{figure}[tb]
    \centering
    \begin{subfigure}{1\linewidth}
        \includegraphics[width=1\linewidth]{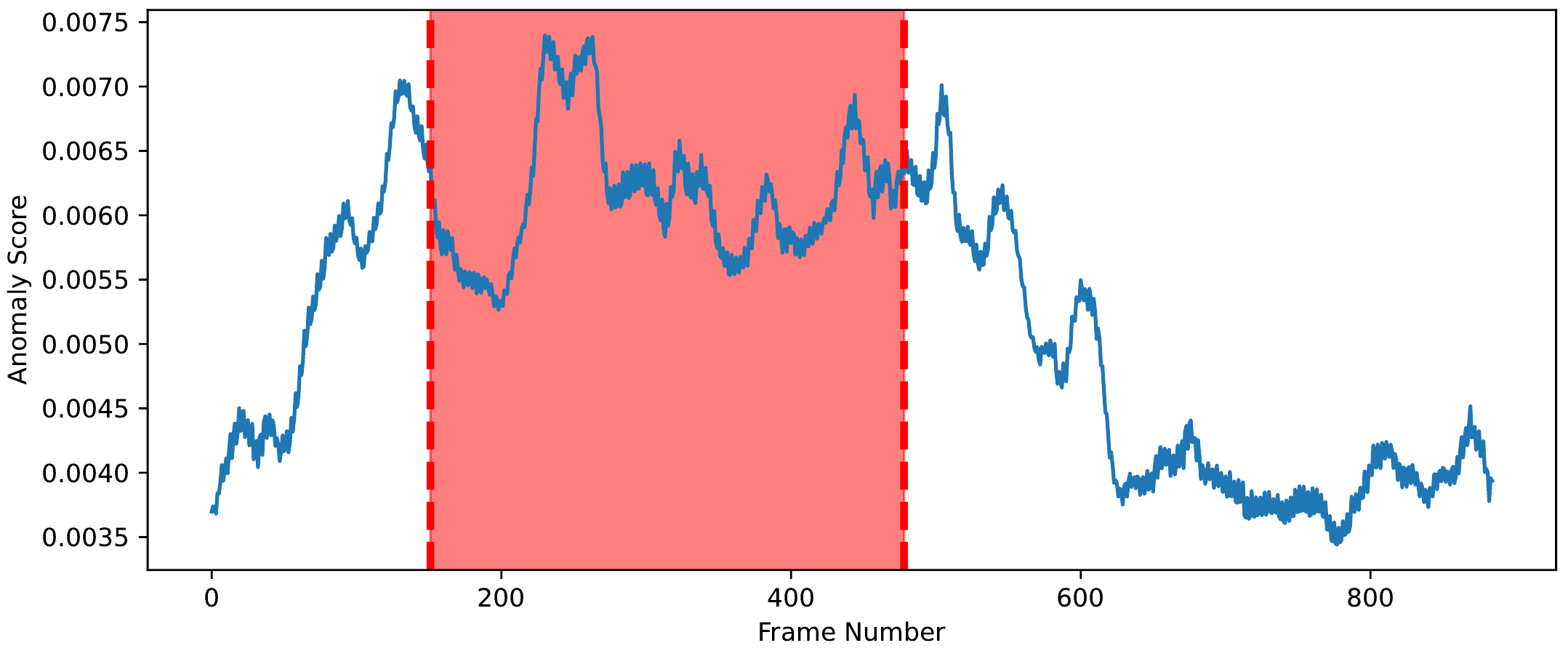}
        \caption{Anomaly scores for sequence A0405 (crossing people start to argue violently, \ie aggressive behavior).}
    \end{subfigure}
    \begin{subfigure}{1\linewidth}
        \includegraphics[width=\linewidth]{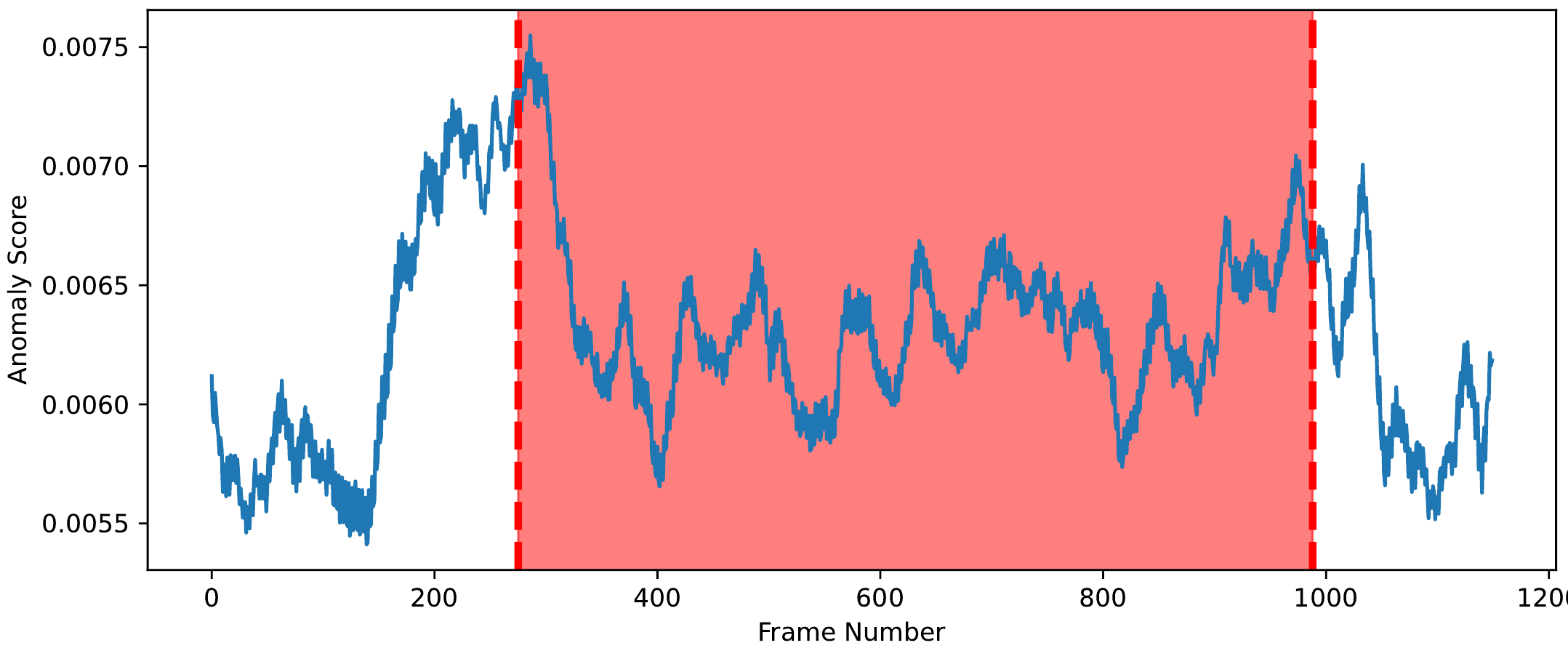}
        \caption{Anomaly scores for sequence A0444 (person lies down on floor, \ie a potential medical issue).}
    \end{subfigure}
    \caption{Examples of the anomaly score generated by the P-CAE network with W-MSE loss. The ground truth anomalous part of the sequence is highlighted in red.}
    \label{fig:ano_score_examples}
\end{figure}

\subsection{Analysis of Anomaly Categories} \label{sec:anom_cats}
Explainability of deep learning methods' outputs is not straightforward and it is thus challenging to analyze why certain anomalies are being detected while others might not in our case. To approach this issue, we split the anomaly types present in the TIMo dataset into three categories and evaluated the performance separately. These categories are:
\begin{itemize}
    \item \textbf{Aggressive behavior}, such as arguing violently and throwing objects.
    \item \textbf{(Potential) Medical Issue}, indicated by people collapsing or staggering \etc
    \item \textbf{Left-behind objects}, \eg a suit case being abandoned by a person walking past.
\end{itemize}

The list of anomaly types assigned to each of the categories is given in the supplementary material along with a full evaluation on each individual anomaly type. The results on the three anomaly categories are shown in \Cref{tab:anomaly_group_results} and reveal major differences in how well the networks are able to detect each of them. All networks are able to detect aggressive behavior reasonably well in the tilted view data, but practically fail mostly at detecting anomalies that indicate medical issues. \Cref{fig:ano_score_examples} provides examples of the anomaly score for a sequence of each case. Results for the top-down view data behave similarly \wrt aggressive behavior, but detection of medical issues is somewhat better. Left-behind objects appear in turn to be less likely to cause high anomaly scores in the top-down view data than in tilted view.

\section{Conclusion} \label{sec:conc}
We presented methods for performing video anomaly detection on depth images from a time-of-flight camera in a fully unsupervised way. To the best of our knowledge, we are the first ones to address this problem. Our results suggest that foreground masks can function as valuable auxiliary data for depth video anomaly detection, similar to the role that optical flow plays for RGB VAD. The integration of foreground masks into the loss function used to train the networks we proposed consistently improved the anomaly detection performance. We also found that autoencoders based on the vision transformer (ViT) architecture are a viable alternative to convolution-based ones. Using a categorization of different anomaly types in the TIMo dataset, we moreover found significant differences in how well each of these categories is detected across all the proposed methods.

The performance of current VAD approaches is mostly still too low for them to replace humans for safety-critical monitoring tasks, but these systems can already be helpful by drawing the attention of a CCTV operator towards potentially relevant events.

Research towards anomaly detection on depth video can make use of many concepts that are already being used on RGB data, but we show that specific strengths of depth data should be leveraged to achieve better performance than by merely carrying over the algorithms for RGB VAD. Depth from ToF sensors has shown to be a promising data modality in combination with autoencoder networks for unsupervised video anomaly detection. It outperformed infrared amplitude in a direct comparison and can help to implement privacy-preserving monitoring systems in the future.

\textbf{Acknowledgements:} This work was partially funded within the Electronic Components and Systems for European Leadership (ECSEL) Joint Undertaking in collaboration with the European Union’s H2020 Framework Program and Federal Ministry of Education and Research of the Federal Republic of Germany (BMBF), under grant agreement 16ESE0424 / GA826600 (VIZTA).

{\small
\bibliographystyle{ieee_fullname}
\bibliography{egbib}
}

\end{document}